\documentclass[11pt,a4paper]{article}
\usepackage[hyperref]{AACL-IJCNLP2020}
\usepackage{times}
\usepackage{latexsym}

\usepackage{amsmath}

\usepackage{url}
\usepackage{graphicx}
\usepackage{booktabs}       
\usepackage{algorithm}
\usepackage{algpseudocode}
\usepackage[disable]{todonotes}
\newcommand{\comment}[1]{}

\usepackage[font=small,labelfont=bf,tableposition=top]{caption}
\usepackage[font=footnotesize]{subfig}
\usepackage{microtype}

\aclfinalcopy 

\newcommand{\mf}[1]{\multicolumn{2}{|c|}{\bf #1}}

\title{Multilingual Translation with Extensible Multilingual Pretraining and Finetuning}

\author{Yuqing Tang, Chau Tran, Xian Li, Peng-Jen Chen, Naman Goyal,\\
{\bf Vishrav Chaudhary, Jiatao Gu, Angela Fan} \\
Facebook AI \\
  \texttt{\{yuqtang,chau,xianl,pipibjc,naman\}@fb.com} \\
  \texttt{\{vishrav,jgu,angelafan\}@fb.com} \\
  }

\date{}

\begin{document}
\maketitle

\begin{abstract}
Recent work demonstrates the potential of multilingual pretraining of creating one model that can be used for various tasks in different languages. Previous work in multilingual pretraining has demonstrated that machine translation systems can be created by finetuning on bitext. In this work, we show that multilingual translation models can be created through \emph{multilingual finetuning}. Instead of finetuning on one direction, a pretrained model is finetuned on many directions at the same time. Compared to multilingual models trained from scratch, starting from pretrained models incorporates the benefits of large quantities of unlabeled monolingual data, which is particularly important for low resource languages where bitext is not available. We demonstrate that pretrained models can be extended to incorporate additional languages without loss of performance. We double the number of languages in mBART to support multilingual machine translation models of 50 languages. Finally, we create the \textsc{ML50} benchmark, covering low, mid, and high resource languages, to facilitate reproducible research by standardizing training and evaluation data. On \textsc{ML50}, we demonstrate that multilingual finetuning improves on average 1 BLEU over the strongest baselines (being either multilingual from scratch or bilingual finetuning) while improving 9.3 BLEU  on average over bilingual baselines from scratch. 
\end{abstract}

\section{Introduction}

A multitude of datasets and models have been developed in natural language processing for a wide variety of tasks and applications. However, a large proportion of these have focused on English. Many works have contributed resources for other languages, developing specialized models for each language of interest is not scalable, not to mention difficult for low resource languages where labeled data is exceptionally scarce. 

Recent work in multilingual NLP shows promise for incorporating many languages into one architecture. For example, the mBART~\cite{liu2020multilingual} model trains on twenty five different languages and can be finetuned for various different tasks. For translation, mBART was finetuned on bitext (bilingual finetuning). 
However, while mBART was trained on a variety of languages, the multilingual nature of the pretraining is not used during  finetuning. Finetuning on bitext to translate from one language to another does not leverage the full capacity of the multilingual pretraining. Instead, we propose \emph{multilingual finetuning} of pretrained models, and we demonstrate large improvements compared to bilingual finetuning. 

Previous work ~\cite{aharoni-etal-2019-massively,arivazhagan2019massively,zhang2020improving} has explored multilingual translation by training multiple directions within the same model from scratch, but this approach faces challenges for mid to low resource languages. In  lower resource scenarios, bitext data is usually unavailable in large quantities, making it challenging to train from scratch. In contrast, monolingual data exists even for low resource languages, particularly in resources such as Wikipedia or Commoncrawl, a version of the web. Thus, leveraging this monolingual data through pretraining can provide a much stronger starting point for low resource machine translation tasks.

However, unlike training a multilingual model from scratch, pretrained models are limited to the choices made during pretraining. For example, mBART was only trained on 25 languages, so finetuning to translate on a model not part of these 25 languages is not possible. Thus, people are restricted to the languages selected to train the initial model, as it is incredibly computationally intensive to retrain from scratch. In this work, we show that existing pretrained models, such as mBART ~\cite{liu2020multilingual} can be extended to  additional languages. We demonstrate by \emph{doubling} the number of languages supported by mBART --- to 50 --- without loss of performance on the original 25 languages and without starting from scratch. This allows languages to be added flexibly, while preserving the broader utility of the pretrained model, as it can be used for tasks beyond translation. 

Further, working in a multilingual setting remains challenging, as various different datasets, evaluation settings, and preprocessing such as tokenization are used. Benchmarks for sentence embeddings~\cite{hu2020xtreme}, natural language inference ~\cite{conneau2018xnli}, and question answering ~\cite{lewis2019mlqa} exist, but there is not yet a setting for machine translation. To this end, we contribute the \textsc{ML50} benchmark, a dataset of 50 languages with publicly available training and evaluation sets, including high, mid, and extremely low resource directions. We will open source this benchmark for the community. 

We make three main contributions:
\begin{itemize}
\itemsep0em 
    \item An effective and novel approach for multilingual translation models with multilingual pretraining (with monolingual data) followed by multilingual finetuning (with parallel data).  In the Many-to-English setting, multilingual finetuning achieves a 3.6 BLEU improvement over bilingual finetuning, and 2.6 BLEU improvement compared to multilingual models trained from scratch. On average, combining Many-to-English and English-to-Many, multilingual finetuning improves $1$ BLEU points over the strongest baseline.
    

    \item We show that existing pretrained models, such as mBART, can be extended to incorporate additional languages without training from scratch and without performance loss on the original languages. We release \emph{mBART50} for the community to use, which has double the number of languages of the original mBART. 
    \item To facilitate reproducible research on multilingual translation with representative challenges of the real world, we create the \textsc{ML50} benchmark covering high, mid, and low resource languages and consisting of 230M bitext. 
\end{itemize}

\section{Related work}
\subsection{Multilingual Denoising Pretraining}
This work is related to recent progress of pretraining techniques for NLP applications~\cite{peters2018deep,radford2018gpt,devlin2018bert,liu2019roberta,song2019mass,lewis2019bart}. In particular, recent works explored pre-training on multilingual unlabeled corpus~\cite{lample2019cross,conneau2019unsupervised,liu2020multilingual,tran2020cross}, and significantly improved the performance of fine-tuning on machine translation between two languages. We extend \newcite{liu2020multilingual} by allowing fine-tuning in multilingual settings.

\subsection{Multilingual Neural Machine Translation} 
Training a universal translation system between multiple languages~\cite{firat2016multi,johnson2017google} has shown enormous improvement for translating low-resource languages~\cite{gu-etal-2018-universal}, and even enabling zero-shot translation~\cite{gu2019improved,arivazhagan2019missing}. \newcite{arivazhagan2019massively} indicates that it is essential to train gigantic models with enough capacity to fully leverage massive multilingual corpora. 

A closely related concurrent work, \newcite{siddhant2020leveraging} shows it is possible to train a multilingual system jointly with monolingual datasets based on \newcite{song2019mass}. It naturally enables translation for languages without parallel data. In contrast, this work focuses on fine-tuning multilingual translation systems given a pre-trained model.

\section{Multilingual Translation from Denoising Pretraining}
We briefly describe the pretrained multilingual BART model and present \emph{multilingual finetuning}, a technique to convert pretrained models into multilingual machine translation systems.

\paragraph{mBART} multilingual BART (mBART)~\cite{liu2020multilingual} is a sequence-to-sequence generative pretraining scheme. The model incorporates $N$ languages by concatenating data: 
$\mathcal{D}=\{\mathcal{D}_1, ..., \mathcal{D}_N \}$ where each $\mathcal{D}_i$ is a collection of monolingual documents in language $i$. mBART is trained as a denoising autoencoder, training to predict the original text $X$ given $g(X)$ where $g$ is a noising function that corrupts text. We maximize $\mathcal{L}_\theta$: 
\begin{equation}
    \mathcal{L}_\theta = 
    \sum_{\mathcal{D}_i\in \mathcal{D}}
    \sum_{x\in \mathcal{D}_i}\log P(x | g(x); \theta)~,
    \label{eq:learning}
\end{equation}
where $x$ is an instance in language $i$ and the distribution $P$ is defined by the seq-to-seq model. This model is pretrained using two types of noise in $g$ --- random span masking and order permutation --- as described in \cite{liu2020multilingual}.

\subsection{Multilingual Finetuning}
To leverage multilingual pretraining to create translation systems, previous work~\cite{liu2020multilingual} used mBART as a starting point and then performed bilingual finetuning. Concretely, the seq-to-seq model was finetuned on language $i$ to language $j$ translation. However, \emph{bilingual finetuning} does not leverage the full capacity of multilingual pretraining. Recent work on multilingual translation~\cite{aharoni-etal-2019-massively,arivazhagan2019massively} displays that strong translation models can be created by doing multilingual training rather than using bilingual tranining. Instead of training a model from language $i$ to language $j$, a model is trained to translate N languages to N other languages. 

Thus, we propose to do \emph{multilingual finetuning} (\textbf{ML-FT}) to adapt pretrained models to become multilingual models. This procedure creates one model capable of translating many languages to many other languages, which has efficiency and storage maintenance benefits. Further, multilingual finetuning retains several benefits of multilingual translation models in general, for example allowing languages of similar family to benefit each other.

To perform multilingual finetuning, we collect bitexts of different language pairs $(i, j)$ into a collection  $\mathcal{B}_{i,j} = \{ (x_i, y_j)  \}$ for each direction $(i,j)$. Following mBART~\cite{liu2020multilingual}, we augment each bitext pair $(x_i, y_j)$ by adding a source language token and a target language token at the beginning of $x$ and $y$ respectively to form a target language token augmented pair $(x', y')$. We then initialize a transformer based seq-to-seq model by the pretained mBART, and provide the multilingual bitexts $\mathcal{B}=\bigcup_{i,j} \mathcal{B}_{i,j}$ to finetune the pretrained model. 

\paragraph{Multilingual Translation Model Variants} We explore $3$ configurations to create different versions of multilingual translation models: \emph{Many-to-one} ($N \rightarrow 1$), \emph{one-to-Many} ($1 \rightarrow  N$), and \emph{Many-to-Many} ($N \leftrightarrow N$) via a pivot language. Concretely, the Many-to-one model encodes $N$ languages and decodes to English, while the one-to-Many model encodes English and decodes into $N$ languages. Finally, the Many-to-Many model encodes and decodes $N$ languages. We follow~\cite{arivazhagan2019massively} and use pivot data through English to create Many-to-Many models. 

\paragraph{Temperature Sampling} When training multilingual models with many languages, the training dataset sizes are imbalanced as different languages have different quantities of bitext. Thus, we train with temperature upsampling, which upsamples lower resource pairs so that the high resource languages do not dominate the training data. We follow~\citet{arivazhagan2019massively} and use the following temperature based sampling function with temperature $T$ to sample data for each direction:
\begin{align*}
  p_{i, j} &\propto \left (\frac{|\mathcal{B}_{i,j}|}{\sum_{i,j} |\mathcal{B}_{i,j}| } \right)^{1/T}
\end{align*}

\section{Results from Multilingual Finetuning on $25$ Languages}

We first examine the impact of multilingual finetuning directly on existing pretrained models. We present results on the 25 languages included in mBART, using the existing mBART model. First, we describe three strong baselines: bilingual finetuning, bilingual translation models from scratch, and multilingual translation models from scratch. Then, we describe our experimental setting. Finally, we present results on 25 languages, showing that on average, multilingual finetuning improves $0.2$ BLEU over the strongest baseline --- 1.0 BLEU point improvement over the strongest to-English baseline while $-0.63$ difference to the strongest from-English baseline. 

\begin{table*}[ht]
    \centering
    \begin{tabular}{|c|c|c|c|c|c|c|c|c|c|c|c|}
    \toprule
    \textbf{Data} &  \multicolumn{5}{|c|}{\textbf{Translation to English}} & \multicolumn{5}{|c|}{\textbf{Translation from English}} \\   
     \cline{2-11}
    & \bf BL-FT &  \mf{ML-Scratch} &  \mf{ML-FT} & \bf BL-FT& \mf{ML-Scratch} &  \mf{ML-FT}    \\     
     &  $\to$en &   N$\to$1  &  N$\leftrightarrow$N &  N$\to$1  &  N$\leftrightarrow$N & en$\to$ & 1$\to$N  &  N$\leftrightarrow$N &  1$\to$N  &  N$\leftrightarrow$N \\    
    \midrule
$>$10M & 2.60 & 3.99 & 2.51 &    \bf 4.37  &  3.19 & \bf 1.67   & 0.64 & -0.6 & 2.20 & -0.90 \\
1M-10M & 3.70 & 5.70 & 5.06 &    \bf  6.40 &  4.62 & \bf  3.40 &  2.40 & 1.7 & 1.76 & 1.40 \\
100k-1M & 5.49 & 7.28 & 7.04 &   \bf  8.13 &  6.47 &     4.17 &  \bf  4.31 & 4.97 & 2.9 & 2.00 \\
7k-30k & 10.80 & 14.63 & 13.77 & \bf 18.03 &  14.57 &   7.27 & \bf    8.07 & 7.90 & 7.6 & 0.90 \\
\hline
All &    4.94 & 6.91 & 6.15 &    \bf  7.91 &  6.14 & \bf   3.67 &    3.31 & 2.66 & 3.0 & 1.81 \\
    \bottomrule
\end{tabular}
    \vspace{.2cm}
    \caption{\textbf{Multilingual Finetuning on $25$ languages comparing to bilingual models}. Numbers are the improvement in BLEU compared to bilingual training from scratch.}
    \label{tab:main25-quality}
\end{table*}

\begin{table*}[ht]
    \centering
    \begin{tabular}{|c|c|c|c|c|c|c|c|c|c|}
    \toprule
    \textbf{Data} &  \multicolumn{4}{|c|}{\textbf{Translation to English}} & \multicolumn{4}{|c|}{\textbf{Translation from English}} \\   
    \cline{2-9}
      & \mf{ML-FT vs BL-FT} &  \mf{ML-FT vs ML-SC} &  \mf{ML-FT vs BL-FT} &  \mf{ML-FT vs ML-SC}    \\     
      &   N$\to$1  &  N$\leftrightarrow$N &  N$\to$1  &  N$\leftrightarrow$N &   1$\to$N  &  N$\leftrightarrow$N &  1$\to$N  &  N$\leftrightarrow$N\\ 
    \midrule
$>$10M & 1.77 & 0.59 & 0.39 & -0.80 & 0.53 & -0.64 & 1.56 & 1.61 \\
1M-10M & 2.70 & 0.92 & 0.70 & -1.08 & -1.64 & -2.68 & -0.64 & -1.00 \\
100k-1M & 2.64 & 0.98 & 0.86 & -0.81 & -1.29 & -2.64 & -1.43 & -2.44 \\
7k-30k & 7.23 & 3.77 & 3.40 & -0.07 & 0.33 & -0.93 & -0.47 & -1.57 \\
\hline
All & 2.98 & 1.20 & 1.00 & -0.77 & -0.63 & -1.85 & -0.28 & -0.85 \\
    \bottomrule
\end{tabular}
    \vspace{.2cm}
    \caption{\textbf{Multilingual Finetuning on $25$ languages comparing to bilingual finetuning and multilingual training from scratch}. Numbers are the improvement in BLEU compared to bilingual finetuning and multilingual training from scratch.  We compare to bilingual finetuning (BL-FT) and multilingual translation from scratch (ML-SC). We perform multilingual finetuning on the existing mBART model. On average, multilingual finetuning (ML-FT) improves $1.0$ BLEU in Many-to-one (N$\rightarrow$1), $-0.77$ BLEU in one-to-Many (1$\rightarrow$N), and $-0.77$ and $-1.85$ BLEU for to-English and from-English respectively in Many-to-Many (N$\leftrightarrow$N) settings compared to the strongest baselines ML-SC many-to-one, BL-FT, and ML-SC many-to-one and BL-FT finetuning (combined baselines for ML-FT many-to-many) respectively.}
    \label{tab:mbt-quality-25}
\end{table*}

\subsection{Baselines}
We compare our proposed multilingual finetuning to three strong baselines: bilingual training from scratch, bilingual finetuning, and multilingual models trained from scratch.

\paragraph{Bilingual Trained from Scratch (BL-Scratch)}
We train bilingual translation models with standard Transformer~\cite{vaswani2017attention} models~\footnote{5 layers with 512 embedding dimension, 2048 FFN embedding dimension, and 8 heads for both encoder and decoder} for translation into and from English to $49$ languages. For directions with more than 1 million bitext training data (de, cs, fr, ja, es, ru, pl, zh, fi, lv, lt, and hi ), we train Transformer Big models~\footnote{6 layers  with 1024 embedding dimension, 4096 FFN embedding dimension, and 16 heads for both encoder and decoder} as there is more data to benefit from additional model capacity. For directions with more than 10 million bitext training data (de, cs, fr, ja, es, ru, pl, and zh), we train Transformer Large models~\footnote{12 layers  with 1024 embedding dimension, 4096 FFN embedding dimension, and 16 heads for both encoder and decoder} as there is even more data to benefit from additional model capacity. The best performing bilingual model is selected as the Bilingual Train from Scratch baseline. 

\paragraph{Bilingual Finetuning (BL-FT)} 
Bilingual finetuning adapts the mBART model into bilingual machine translation models by training for longer on translation bitext. For each language direction, we follow~\citet{liu2020multilingual} and finetune for $40$K updates to obtain the Bilingual Finetuning baseline. 

\paragraph{Multilingual Trained from Scratch (ML-SC)}
We train $3$ different multlilingual models from scratch: Many-to-one (N$\rightarrow$1), one-to-Many (1$\rightarrow$N), and Many-to-Many (N$\leftrightarrow$N) with English as pivot. We train for $500$K updates and sweep through different batch sizes, learning rates, and upsampling temperature for best performing multilingual model on validation, using $32$ GPUs for each training instance. Following~\citet{arivazhagan2019massively}, we train with temperature upsampling.



\subsection{Evaluation and Generation} 
We evaluate performance with tokenized BLEU, following the tokenization in mBART \cite{liu2020multilingual}.  To generate, we decode using beam search with beam size $N=5$ with length penalty$=1.0$ on the validation set. We do not perform checkpoint averaging. To select the best performing model in a sweep, we compare BLEU on the validation set.

\subsection{Performance on $25$ Languages}

We first evaluate our proposed multilingual finetuning technique on $25$ languages using the existing mBART model. We compare bilingual finetuning from mBART (BL-FT), multilingual training from scratch (ML-SC), and multilingual finetuning (ML-FT) by quantifying the BLEU improvement over the bilingual training from scratch baseline. Results are displayed in Table~\ref{tab:main25-quality}, separated into three settings: Many-to-one (N$\rightarrow$1), one-to-Many (1$\rightarrow$N), and Many-to-Many (N$\leftrightarrow$N).

\paragraph{Performance of Multilingual Finetuning} Compared to the BL-FT and ML-SC baselines, multilingual finetuning has consistently stronger results in the Many-to-one setting, translating from 25 different languages into English. The improvement is 7.9 BLEU points stronger than the bilingual from scratch baseline, and 1.0 BLEU points stronger than the the strongest baseline, ML-SC.  

However, in the one-to-Many setting, improvement of all multilingual methods against bilingual baselines is lower across the board. We hypothesize this is due to the challenge of needing to decode into many different languages (additional analysis is presented in Section~\ref{sec:one-to-many-challenge}). Multilingual finetuning method is $3$ BLEU points stronger than the bilingual from scratch baseline; it is also comparable to the strongest baseline --- bilingual finetuning with $-0.6$ BLEU difference on average.

Finally, in the Many-to-Many setting, improvement of all many-to-many multilingual methods against bilingual baselines is lower across the board. Again we hypothesize this is due to the challenge of decoding into many different languages including English (additional analysis is presented in Section~\ref{sec:one-to-many-challenge}). Multilingual finetuning method is $3.98$ BLEU points stronger than the bilingual from scratch baseline for translation from and into English combined. Overall, it is lower than the strongest from-English and into-English baselines combined with $-1.3$ BLEU difference on average.

\paragraph{Performance by Resource Level}

Comparing the languages by resource level, we see that the improvement from multilingual training is more significant as the quantity of training bitext decreases. For example, in the multilingual finetuning (ML-FT) Many-to-one setting, improvement over bilingual from scratch is 4.4 BLEU points for languages with more than 10M bitext, but is 18.0 BLEU points for languages with 7K-30K available bitext. The trend is less consistent in the one-to-Many setting, but low resource languages still see improvements. For example, with multilingual finetuning (ML-FT), improvement over bilingual from scratch is 2.2 BLEU for languages with more than 10M bitext, but 7.6 BLEU for languages with 7K-30K available bitext.

\begin{table*}[ht]
    \centering
    \begin{tabular}{|r|p{0.8\textwidth}|}
    \toprule
    \textbf{Data size} & \textbf{Languages} \\ 
    \midrule 
    \bf 10M+ & German, Czech, French, Japanese, Spanish, Russian, Polish, Chinese \\ 
    \hline
    \bf 1M - 10M & Finnish, Latvian, Lithuanian, Hindi, Estonian  \\
    \hline
    \bf 100k to 1M & Tamil, Romanian, Pashto, Sinhala, Malayalam, Dutch, Nepali, Italian, Arabic, Korean, Hebrew, Turkish, Khmer, Farsi, Vietnamese, Croatian, Ukrainian\\ 
    \hline
    \bf 10K to 100K & Thai, Indonesian, Swedish, Portuguese, Xhosa, Afrikaans, Kazakh, Urdu,  Macedonian, Telugu, Slovenian, Burmese, Georgia    \\
    \hline
    \bf 10K- & Marathi, Gujarati, Mongolian, Azerbaijani, Bengali\\ 
    \bottomrule
    \end{tabular}
    \vspace{0.2cm} 
    \caption{\textbf{Languages in \textsc{ML50} Benchmark.} We display the languages included in the \textsc{ML50} Benchmark and the quantity of training data in bitext pairs. Full breakdown is provided in Appendix Table~\ref{tab:ml50_dataset}.}
    \label{tab:ml50}
\end{table*}

\section{Results from Multilingual Finetuning on $50$ Languages}

Multilingual finetuning showed strong improvements on $25$ languages in the Many-to-one setting and we subsequently extend to incorporate a greater number of languages --- 50 instead of 25. However, the number of languages possible is limited by the initial selection of languages in mBART. To remedy this, we show that the number of languages in mBART can be easily extended with additional pretraining. Second, we build the \textsc{ML50} benchmark, to standardize training data, evaluation data, and evaluation procedure across 50 different languages. Finally, we display results of multilingual finetuning from mBART on 50 languages and show strong improvements over the baselines.

\subsection{Doubling the Languages in mBART}

We describe how we extend existing pretrained models to incorporate a greater number of languages. This technique allows existing models to be used on new languages, rather than needing to restart a computationally intensive pretraining method from scratch. 

\paragraph{Creating mBART50} While multilingual pretrained models have shown strong performance in a variety of tasks ~\cite{liu2020multilingual,conneau2019unsupervised}, they remain limited as they are trained on a fixed number of languages. For example, mBART was trained on 25 languages, all fairly high resource. Pretraining fully from scratch is computationally intensive --- mBART trained for 2.5 weeks on 256 Nvidia V100 GPUs \cite{liu2020multilingual}. However, there are hundreds of different languages in the world, so restarting pretraining from scratch to add any of them to mBART would be difficult. Instead, we take the existing mBART model, trained on $25$ languages, and show that it can be extend to more than $50$ languages. We take the public available pretrained  mBART model\footnote{\url{https://github.com/pytorch/fairseq/tree/master/examples/mbart}} which was pretrained on $25$ languages and extend its embedding layers with randomly initialized vectors for an extra set of $25$ language tokens. We then combine the monolingual data of original $25$ languages and the new $25$ languages together to continue pretraining this extended MBART model. We will release the mBART50 model as a general purpose multilingual pretrained model, which will be useful for a variety of generation tasks beyond machine translation.

\paragraph{Data and Training Details} We use the \texttt{mBART.cc25} checkpoint~\cite{liu2020multilingual} available in the \texttt{fairseq} library \cite{ott2019fairseq} to continue the pretraining process. We use the monolingual data from XLMR~\cite{conneau2019unsupervised} to extend the pretraining to a set of $25$ languages in addition to the $25$ languages mBART model. To be consistent mBART, we reuse its $250$K sentencepiece~\cite{kudo-richardson-2018-sentencepiece} model which was trained using monolingual data for $100$ languages from XLMR, and thus already supports languages beyond the original 25 mBART was trained on. 
For pre-training, we train mBART50 for an additional $300$K updates with a batch size of $1700$ tokens. The sizes of the monolingual data for the additional 50 languages is provided in the appendix. 

\begin{table*}[ht]
    \centering
    \begin{tabular}{|c|c|c|c|c|c|c|c|c|c|c|c|}
    \toprule
    \textbf{Data} &  \multicolumn{5}{|c|}{\textbf{Translation to English}} & \multicolumn{5}{|c|}{\textbf{Translation from English}} \\   
    \cline{2-11}
    & \bf BL-FT &  \mf{ML-SC} &  \mf{ML-FT} & \bf BL-FT& \mf{ML-SC} &  \mf{ML-FT}    \\     
     &  $\to$en &   N$\to$1  &  N$\leftrightarrow$N &  N$\to$1  &  N$\leftrightarrow$N & en$\to$ & 1$\to$N  &  N$\leftrightarrow$N &  1$\to$N  &  N$\leftrightarrow$N\\ 
    \midrule
$>$10M & 2.7 & 2.8 & 1.9 & \bf 3.8 & 1.4 & \bf 1.9 & -0.6 & -1.7 & -0.3 & -1.7 \\
1M-10M & 3.9 & 4.8 & 4.1 & \bf 6.2 & 4.4 & \bf 3.3 & 1.5 & 1.0 & 1.7 & 0.6 \\
100k-1M & 5.7 & 6.9 & 7.0 & \bf 8.2 & 7.4 & \bf 4.4 & 4.0 & 3.4 & 4.0 & 3.2 \\
10K-100K & 16.8 & 17.9 & 18.3 & \bf 22.3 & 20.6 & 13.4 & 13.6 & \bf 13.9 & 13.5 & 13.6 \\
4k-10k & 11.6 & 13.1 & 14.1 & 1\bf 8.9 & 15.0 & 8.7 & 10.6 & \bf 10.9 & 9.9 & 9.7 \\
\hline
All    & 8.7 & 9.7 &  9.8 & \bf 12.3 & 10.6 & \bf 6.8 & 6.4 & 6.0 & 6.3 & 5.7 \\
    \bottomrule
\end{tabular}
    \vspace{.2cm}
    \caption{\textbf{Multilingual Finetuning on $50$ languages comparing to bilingual models}. Improvement in BLEU compared to bilingual training from scratch is shown. }
    \label{tab:main-quality-50}
\end{table*}

\begin{table*}[ht]
    \centering
    \begin{tabular}{|c|c|c|c|c|c|c|c|c|c|}
    \toprule
    \textbf{Data} &  \multicolumn{4}{|c|}{\textbf{Translation to English}} & \multicolumn{4}{|c|}{\textbf{Translation from English}} \\   
    \cline{2-9}
      & \mf{ML-FT vs BL-FT} &  \mf{ML-FT vs ML-SC} &  \mf{ML-FT vs BL-FT} &  \mf{ML-FT vs ML-SC}    \\     
      &   N$\to$1  &  N$\leftrightarrow$N &  N$\to$1  &  N$\leftrightarrow$N &   1$\to$N  &  N$\leftrightarrow$N &  1$\to$N  &  N$\leftrightarrow$N\\ 
    \midrule
$>$10M & 1.05 & -1.34 & 0.95 & -0.50 & -2.15 & -3.53 & 0.31 & -0.01 \\
1M-10M & 2.34 & 0.54 & 1.36 & 0.30 & -1.60 & -2.74 & 0.18 & -0.44 \\
100k-1M & 2.43 & 1.68 & 1.28 & 0.36 & -0.36 & -1.21 & 0.01 & -0.25 \\
10K-100K & 5.49 & 3.82 & 4.37 & 2.30 & 0.06 & 0.21 & -0.13 & -0.25 \\
4k-10k & 7.33 & 3.42 & 5.83 & 0.87 & 1.27 & 1.00 & -0.65 & -1.20 \\
\hline
All & 3.61 & 1.85 & 2.61 & -0.15 & -0.47 & -1.10 & -0.04 & -0.35 \\
    \bottomrule
\end{tabular}
    \vspace{.2cm}
    \caption{\textbf{Multilingual Finetuning on $50$ languages comparing to bilingual finetuning and multilingual training from scratch} We compare to bilingual finetuning (BL-FT) and multilingual translation from scratch (ML-SC). On average, multilingual finetuning (ML-FT) improves $2.61$ BLEU in Many-to-one (N$\rightarrow$1), $-0.47$ BLEU in one-to-Many (1$\rightarrow$N), and $-0.15$ and $-0.35$ BLEU for to-English and from-English respectively in Many-to-Many (N$\leftrightarrow$N) settings compared to the strongest baselines ML-SC many-to-one, BL-FT, and ML-SC many-to-one and BL-FT finetuning (combined baselines for ML-FT many-to-many) respectively.}
    \label{tab:mbt-quality-50}
\end{table*}

\subsection{\textsc{ML50} Benchmark}
\label{sec:dataset}

To demonstrate the impact of multilingual finetuning on additional languages, we create the \textsc{ML50} Benchmark. \textsc{ML50} standardizes the training and evaluation schemes across 50 different languages, from extremely low resource languages like Xhosa and Gujarati to high resource languages like French and German. The full list of languages is shown in Table~\ref{tab:ml50}. We group the languages into five categories based on the amount of available training data: more than 10M pairs (8 languages), 1M to 10M pairs (5 languages), 100k to 1M pairs (17 languages), 10K to 100K pairs (13 languages), and finally, less than 10K pairs of training data (5 languages). \textsc{ML50} includes languages in N language families, from Germanic and Romance languages to Indic and African ones. Many additional languages we contribute are lower resource, compared to the languages in the original mBART. 

\paragraph{Training Data}
We gather parallel data between English and 49 other languages to form \textsc{ML50}, to enable the training of machine translation models. We select these 49 languages based on the amount of parallel and monolingual data to cover languages with different amount of resources and under different language families. The quantity of available monolingual data is relevant for pretraining, so we want to ensure there is a sufficient amount. All of the data is publicly available, such as WMT, IWSLT, WAT, TED, and other published research works. For training data, each language pair can include multiple sources. We simply concatenate them together and remove duplicated source-target sentence pairs for each language pair. We use \texttt{fasttext}~\cite{joulin2017bag} to perform language identification on both source and target sentences, and we remove sentences pairs if either source or target sentence is not predicted as expected language. We further filter out training data that match to any source or target side sentences in evaluation datasets. Compared to other datasets such as \textsc{opus100}, the \textsc{ML50} benchmark contains around 4 times more training data. The full list of languages, data sources, and amount of resulting data can be found in Table~\ref{tab:ml50_dataset} in the Appendix. 

\paragraph{Evaluation Data} 

To ensure high quality evaluation of languages covered in \textsc{ML50}, we include publicly available, widely used evaluation sets. We source these evaluation datasets from translation workshops such as WMT, IWSLT, WAT, and other published research works. We follow the evaluation protocol, including tokenization, used for each of these evaluation sets, to ensure our results are comparable with existing work. We release these scripts to make it easier for others. Compared to other datasets such as \textsc{opus100}, we choose to use high quality existing evaluation datasets rather than use part of the training data as evaluation. This is because training data, particularly for low resource languages, is often very noisy and unreliable. 

\begin{figure*}[ht]
\centering
\includegraphics[width=.47\textwidth]{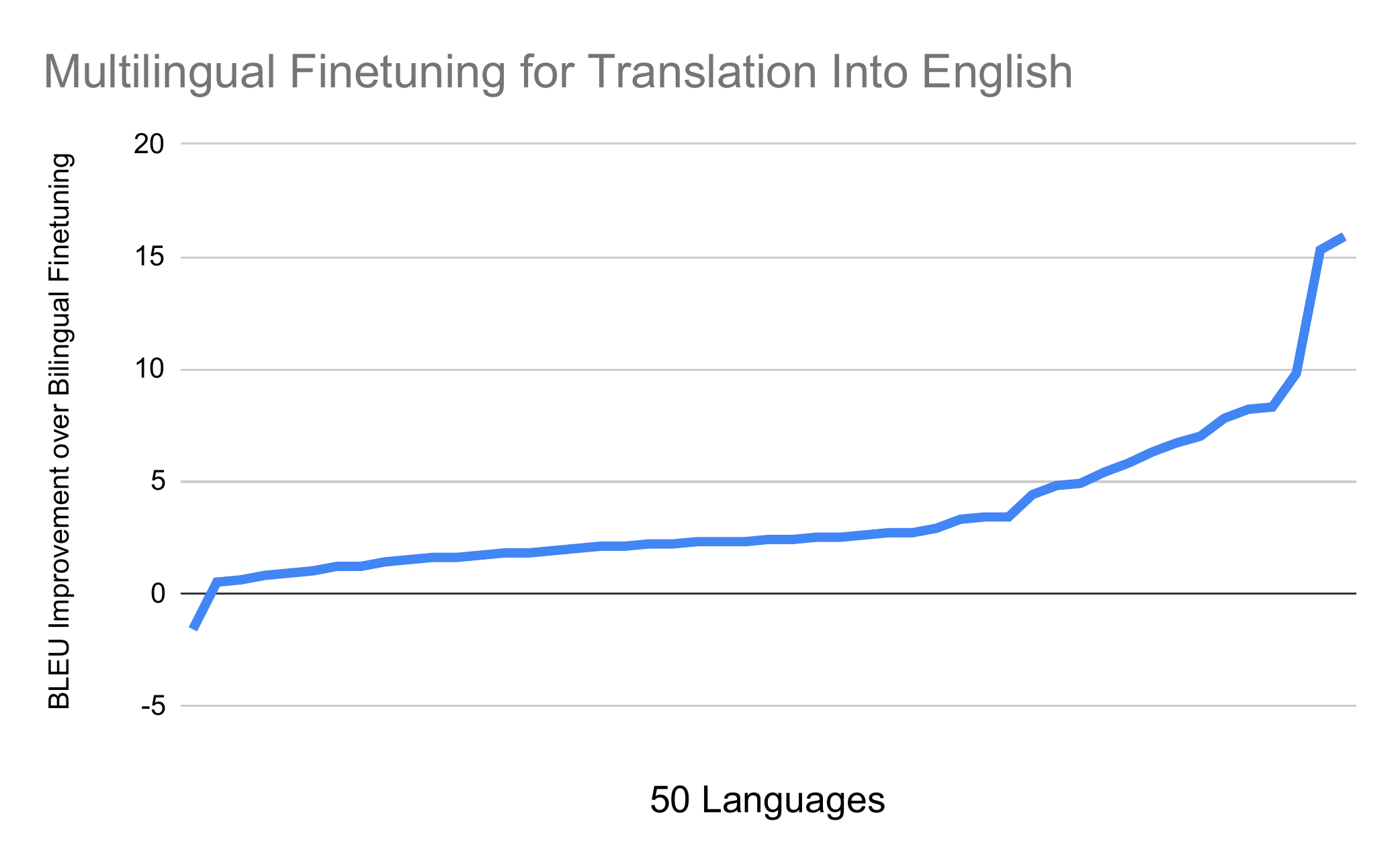}
\includegraphics[width=.47\textwidth]{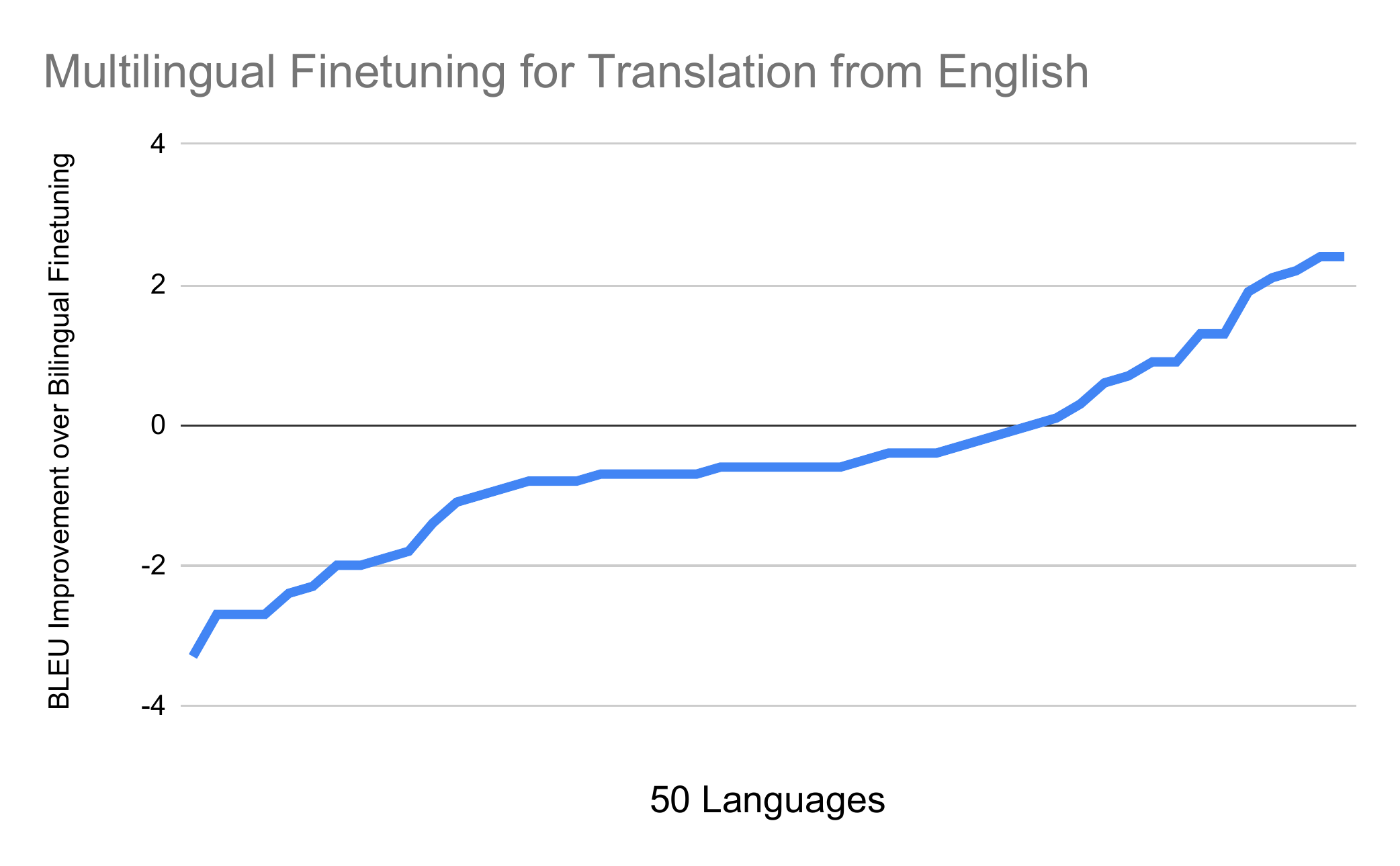}
\caption{Multilingual Finetuning Improvement over Bilingual Finetuning for $50$ Languages Translation: $3.6$ average BLEU improvement for translation into English; $-0.47$ BLEU  average difference for Translation from English}
\label{fig:fig1}
\end{figure*}

\subsection{Performance on 50 Languages}

We evaluate the performance of mBART50 on the \textsc{ML50} Benchmark. We compare to the same baselines --- bilingual finetuning, bilingual training from scratch, and multilingual training from scratch. Results are displayed in Table~\ref{tab:main-quality-50}. 

In the Many-to-One setting averaged across all languages, multilingual finetuning improves over the strongest baseline, multilingual many-to-many from scratch, by 2.5 BLEU points. For lower resource language pairs, the improvement is much more significant. For example, the improvement for languages with 4K-10K training data is 4.8 BLEU points over the strongest baseline, and the improvement for languages with 10K-100K training data is 4+ BLEU over the strongest baseline. 

For One-to-Many, the performance of all methods --- bilingual finetuning, multilingual from scratch, and multilingual finetuning --- is similar. On average, all models have around 5.7 to 7 BLEU points improvement over bilingual baselines. 

Finally, in Many-to-Many, multilingual finetuning achieves 0.8 improvement in the to-English direction over the strongest baseline. In the from-English direction, the performance of Many-to-Many from multilingual finetuning is similar to multilingual from scratch, both around 5.5 to 6 BLEU improvement over bilingual baselines. 

\subsection{Comparison to Bilingual Finetuning}

We examine the performance of our proposed multilingual finetuning method compared to bilingual finetuning. Current work shows that strong translation models can be created by finetuning pretrained models to bilingual translation models. However, this means that a separate model would need to be created for each translation direction of interest, which creates a large quantity of models that need to be finetuned. In contrast, multilingual finetuning allows a multitude of directions to be captured within one model. 

However, multilingual finetuning would mean that the same model capacity must model many directions rather than just one, which could decrease performance. In Figure~\ref{fig:fig1}, we analyze the improvement of multilingual finetuning over the bilingual finetuning. On the left, we compare the Many-to-one setting translating into English, and on the right we compare the one-to-Many setting translating out of English to many different languages. 

In the Many-to-one setting, every language pair except one is improved by multilingual finetuning. Some low resource languages see substantial improvement of 10+ BLEU points, with the largest improvement being over 15 BLEU improvement. On average, multilingual finetuning improves $12.3$ BLEU across all directions into English. In the one-to-Many setting, performance is about the same between multilingual finetuning and bilingual finetuning, with the average improvement at $6.3$ BLEU across all directions out of English comparing to bilingual baselines.


     

\section{Discussion} 

\subsection{Challenges of one-to-Many}
\label{sec:one-to-many-challenge}

In the Many-to-one setting, where models must encode various different languages and decode into English, large improvements are seen when doing multilingual modeling. Previous work has similarly observed this improvement~\cite{arivazhagan2019massively} in multilingual training from scratch, as multilingual modeling increases the quantity of target-side English data seen by the model. For example, compared to bilingual finetuning, our multilingual finetuning model is exposed to English target side data from 50 different language pairs. 

However, in the one-to-Many setting and the Many-to-Many setting, models must decode into 50 different languages. This is a difficult decoding challenge, as a strong conditional language model must be learned for each language. While pretraining exposes the model to monolingual data, the quantity of monolingual data varies for each language. For lower resource languages, such as Gujarati or Xhosa, the quantity of monolingual data available even through online resources such as Commoncrawl, remains limited. Other work~\cite{arivazhagan2019massively} observes similar trends in performance of one-to-Many. 

Overall, we find that multilingual finetuning performs better than any of our assessed baselines --- bilingual training from scratch, bilingual finetuning, and multilingual training from scratch --- when averaged across the Many-to-one and one-to-Many directions. It is important to note that this effect mainly comes from the strong improvement of the Many-to-one setting, and all approaches have similar performance in the one-to-Many setting. 

\subsection{Comparison of mBART50 on 25 Languages}

\begin{figure}[h]
\centering
\includegraphics[width=.47\textwidth]{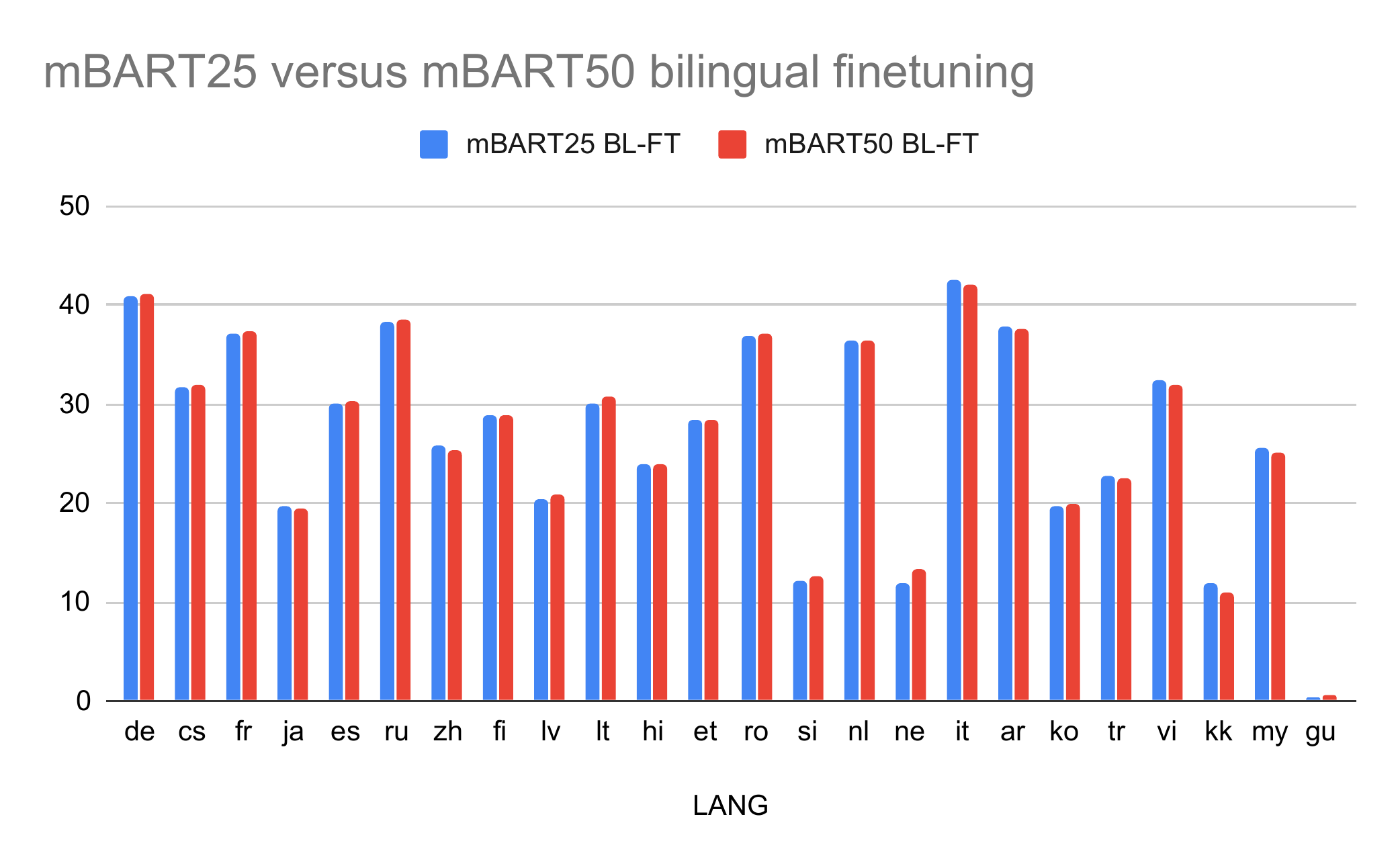}
\caption{Continuing Pretraining with Additional Languages -- No Performance Degeneration in Original Languages}
\label{fig:mbt50-extend}
\end{figure}

We show that the mBART model can be extended from 25 languages to 50 languages without starting from scratch. In this section, we evaluate if adding additional languages is harmful for performance on the original 25 languages. As the model remains the same size but has more to model, it could have reduced capacity for the original 25 languages, but we do not see any reduction in performance. Results are shown in Figure~\ref{fig:mbt50-extend}. For each language, we plot the performance when doing bilingual finetuning with mBART25 and mBART50. We show that performance is almost exactly the same with both models, indicating that the number of languages can be doubled without loss of performance.

\section{Conclusion}

We demonstrate that multilingual neural machine translation models can be created from pretrained models such as mBART. Previous work using pretrained models focused only on bilingual finetuning, and work in multilingual translation trained only from scratch. While using pretrained models could limit the number of languages possible, we show that mBART can be extended to double the number of original languages, without loss of performance on the original languages. We release mBART50 for the community as a strong generative denoising pretrained model in 50 different languages. Further, to train and evaluate on 50 languages, we develop and release the \textsc{ML50} benchmark.
In conclusion, we show that by performing multilingual finetuning, strong improvements of over 2 BLEU points can be achieved in the Many-to-one setting. Overall, averaging across the Many-to-one and one-to-Many directions, our proposed multilingual finetuning strategy outperforms all baselines.

\bibliographystyle{acl_natbib}
\bibliography{ref}
\clearpage

\appendix
\section{Appendices}
\label{sec:appendix}
\begin{table*}[ht]
    \small
    \centering
    \small

\begin{tabular}{|l|r|l|l|r|r|}
\hline
                               & \multicolumn{2}{c|}{ML50 Train}                                 & \multicolumn{3}{c|}{ML50 Eval}                                                                                                                                                                     \\ \hline
\multicolumn{1}{|c|}{Language} & \multicolumn{1}{c|}{\# Sentences} & \multicolumn{1}{c|}{Source} & \multicolumn{1}{c|}{Source} & \multicolumn{1}{c|}{\begin{tabular}[c]{@{}c@{}}\# Sentences\\ Valid\end{tabular}} & \multicolumn{1}{c|}{\begin{tabular}[c]{@{}c@{}}\# Sentences\\ Test\end{tabular}} \\ \hline
af                         & 45967                       & Opus                              & LauraMartinus               & 1500                                                                              & 2686                                                                             \\ \hline
ar                         & 226073                      & IWSLT17                           & IWSLT17                     & 1158                                                                              & 1460                                                                             \\ \hline
az                         & 5680                        & TED58                             & TED58                       & 671                                                                               & 903                                                                              \\ \hline
bn                         & 4487                        & TED58                             & TED58                       & 896                                                                               & 216                                                                              \\ \hline
cs                         & 42587802                    & WMT20                             & WMT19                       & 2983                                                                              & 1997                                                                             \\ \hline
de                         & 45828203                    & WMT20                             & WMT19                       & 2998                                                                              & 2000                                                                             \\ \hline
es *                        & 14524187                    & WMT13                             & WMT13                       & 3003                                                                              & 3000                                                                             \\ \hline
et                         & 1052003                     & WMT18                             & WMT18                       & 2000                                                                              & 2000                                                                             \\ \hline
fa                         & 144895                      & TED58                             & TED58                       & 3930                                                                              & 4490                                                                             \\ \hline
fi *                        & 2353313                     & WMT17                             & WMT17                       & 3000                                                                              & 3002                                                                             \\ \hline
fr                         & 36797950                    & WMT14                             & WMT14                       & 3000                                                                              & 3003                                                                             \\ \hline
gl                         & 9504                        & TED58                             & TED58                       & 682                                                                               & 1007                                                                             \\ \hline
gu                         & 7471                        & WMT19                             & WMT19                       & 1998                                                                              & 1016                                                                             \\ \hline
he                         & 204380                      & TED58                             & TED58                       & 4515                                                                              & 5508                                                                             \\ \hline
hi                         & 1327206                     & ITB                               & ITB                         & 520                                                                               & 2507                                                                             \\ \hline
hr                         & 116792                      & TED58                             & TED58                       & 3333                                                                              & 4881                                                                             \\ \hline
id                         & 83944                       & TED58                             & TED58                       & 2677                                                                              & 3179                                                                             \\ \hline
it                         & 226457                      & IWSLT17.mltlng                    & IWSLT17.mltlng              & 1566                                                                              & 1147                                                                             \\ \hline
ja *                         & 16167141                    & WMT20                             & WMT20 dev-split             & 999                                                                               & 999                                                                              \\ \hline
ka                         & 12364                       & TED58                             & TED58                       & 654                                                                               & 943                                                                              \\ \hline
kk                         & 29186                       & WMT19                             & WMT19                       & 2066                                                                              & 1000                                                                             \\ \hline
km                         & 191967                      & WMT'20                            & Flores devtest              & 2378                                                                              & 2309                                                                             \\ \hline
ko                         & 224612                      & IWSLT17                           & IWSLT17                     & 1143                                                                              & 1429                                                                             \\ \hline
lt *                        & 1395010                     & WMT19                             & WMT19                       & 2000                                                                              & 1000                                                                             \\ \hline
lv *                        & 1808291                     & WMT17                             & WMT17                       & 2003                                                                              & 2001                                                                             \\ \hline
mk                         & 24037                       & TED58                             & TED58                       & 640                                                                               & 438                                                                              \\ \hline
ml                         & 358916                      & lotus                             & lotus                       & 500                                                                               & 1000                                                                             \\ \hline
mn                         & 7168                        & TED58                             & TED58                       & 372                                                                               & 414                                                                              \\ \hline
mr                         & 9397                        & TED58                             & TED58                       & 767                                                                               & 1090                                                                             \\ \hline
my                         & 18073                       & WAT19                             & WAT19                       & 1000                                                                              & 1018                                                                             \\ \hline
ne                         & 227387                      & Flores                            & Flores                      & 2559                                                                              & 2924                                                                             \\ \hline
nl                         & 232572                      & IWSLT17.mltlng                    & IWSLT17.mltlng              & 1777                                                                              & 1181                                                                             \\ \hline
pl                         & 10332683                    & WMT20                             & WMT20 dev-split             & 1000                                                                              & 1000                                                                             \\ \hline
ps                         & 579346                      & WMT'20                            & Flores devtest              & 3162                                                                              & 2698                                                                             \\ \hline
pt                         & 49446                       & TED58                             & TED58                       & 1193                                                                              & 1803                                                                             \\ \hline
ro                         & 592594                      & WMT16                             & WMT17                       & 1999                                                                              & 1999                                                                             \\ \hline
ru *                        & 13922899                    & WMT20                             & WMT19                       & 3000                                                                              & 2000                                                                             \\ \hline
si                         & 565661                      & Flores                            & Flores                      & 2898                                                                              & 2905                                                                             \\ \hline
sl                         & 18751                       & TED58                             & TED59                       & 1068                                                                              & 1251                                                                             \\ \hline
sv                         & 53596                       & TED58                             & TED58                       & 1729                                                                              & 2283                                                                             \\ \hline
ta                         & 609767                      & WMT'20                            & WMT20 dev-split             & 995                                                                               & 994                                                                              \\ \hline
te                         & 22042                       & lotus                             & lotus                       & 500                                                                               & 1000                                                                             \\ \hline
th                         & 93723                       & TED58                             & TED58                       & 2989                                                                              & 3713                                                                             \\ \hline
tr                         & 204200                      & WMT17                             & WMT17                       & 3000                                                                              & 3007                                                                             \\ \hline
uk                         & 104193                      & TED58                             & TED58                       & 3060                                                                              & 3751                                                                             \\ \hline
ur                         & 26302                       & lotus                             & lotus                       & 500                                                                               & 1000                                                                             \\ \hline
vi                         & 127069                      & IWSLT 15                          & IWSLT15                     & 1268                                                                              & 1080                                                                             \\ \hline
xh                         & 48981                       & Opus                              & LauraMartinus               & 1500                                                                              & 2717                                                                             \\ \hline
zh *                        & 10082367                    & WMT20                             & WMT19                       & 3981                                                                              & 2000                                                                             \\ \hline
\end{tabular}

    \caption{ML50 Benchmark dataset stats. For each language, we list the size of training data after the filtering steps, the source of training/evaluation data, and the size of evaluation data. We notice that part of the available dataset are missing due to human error for a few language pairs. We mark these languages with asterisk and we will release next version of the ML50 benchmark data to include the missing data. }
    \label{tab:ml50_dataset}
\end{table*}

\begin{table*}[ht]
    \centering
        \begin{tabular}{| c|c|c|c|c|c|c|c|c|c|c|c|c |}
\toprule
Lang & de & cs & fr & ja & es & ru & pl & zh & fi & lv & lt & hi \\
\midrule
BL-Scratch  to en & 39.7 & 29.0 & 35.2 & 18.4 & 27 & 37.7 & 28.4 & 25.1 & 24.1 & 17.9 & 27.8 & 20.1 \\
BL-FT to en & 41.0 & 32.0 & 37.4 & 19.5 & 30.2 & 38.5 & 31.0 & 25.4 & 28.8 & 20.8 & 30.7 & 23.8 \\
\hline
BL-Scratch  from en & 40 & 24.8 & 39 & 22.2 & 29 & 28.5 & 24.3 & 33.6 & 19.7 & 16.6 & 13.3 & 17.5 \\
BL-FT from en & 41.9 & 26.5 & 40.8 & 24.5 & 30.3 & 30.5 & 26.7 & 35.1 & 23.7 & 19.0 & 16.1 & 20.4 \\
\bottomrule
\toprule
Lang & et & ta & ro & ps & si & ml & nl & ne & it & ar & ko & he \\
\midrule
BL-Scratch  to en & 23.2 & 14.2 & 32.6 & 8.9 & 6.1 & 12.5 & 32.5 & 2.8 & 36.9 & 33.5 & 16.4 & 38.6 \\
BL-FT to en & 28.3 & 18.2 & 37.1 & 15.0 & 12.6 & 18.2 & 36.5 & 13.3 & 42.1 & 37.5 & 19.9 & 42.7 \\
\hline
BL-Scratch  from en & 17.5 & 28.7 & 32.9 & 7.3 & 1.5 & 17.5 & 29.3 & 1.3 & 33.7 & 19.7 & 16.1 & 27.0 \\
BL-FT from en & 22.0 & 34.0 & 37.4 & 9.3 & 4.7 & 25.5 & 33.3 & 6.9 & 38.1 & 22.0 & 20.0 & 29.7 \\
\bottomrule
\toprule
Lang & tr & km & fa & vi & hr & uk & th & id & sv & pt & xh & af \\
\midrule
BL-Scratch  to en & 16.5 & 4.0 & 27.6 & 26.0 & 33.6 & 24.5 & 20.9 & 28.0 & 30.8 & 30.7 & 0.4 & 1.0 \\
BL-FT to en & 22.5 & 8.3 & 33.2 & 31.9 & 42.0 & 33.5 & 28.2 & 36.9 & 44.9 & 46.0 & 12.1 & 26.5 \\
\hline
BL-Scratch  from en & 16.3 & 4.3 & 15.1 & 28.5 & 26.0 & 17.8 & 30.7 & 27.2 & 27.0 & 27.1 & 0.2 & 1.0 \\
BL-FT from en & 22.7 & 5.9 & 18.4 & 32.9 & 32.2 & 24.3 & 36.5 & 35.6 & 38.5 & 41.6 & 11.2 & 18.3 \\
\bottomrule
\toprule
Lang & kk & ur & mk & te & sl & my & ka & gl & mr & gu & mn & az \\
\midrule
BL-Scratch  to en & 1.4 & 7.8 & 14.1 & 10.9 & 7.9 & 3.9 & 6.1 & 6.6 & 2.8 & 0.0 & 3.5 & 2.8 \\
BL-FT to en & 11.0 & 28.0 & 35.8 & 35.8 & 28.5 & 25.1 & 23.8 & 34.3 & 11.6 & 0.5 & 11.2 & 15.5 \\
\hline
BL-Scratch  from en & 0.6 & 8.3 & 8.2 & 15.0 & 4.9 & 19.8 & 3.7 & 4.2 & 5.2 & 0.0 & 3.3 & 1.9 \\
BL-FT from en & 5.9 & 23.7 & 27.2 & 38.8 & 21.9 & 35.8 & 13.0 & 26.7 & 11.5 & 0.6 & 8.5 & 7.4 \\
\bottomrule
 \end{tabular}
    \caption{Bilingual and Finetuning Bilingual Baselines over $50$ languages}
    \label{tab:bl-baselines}
\end{table*}

\begin{table*}[ht]
    \centering
    \begin{tabular}{| c|c|c|c|c|c|c|c|c|c|c|c|c |}
\toprule
Lang & de & cs & fr & ja & es & ru & pl & zh & fi & lv & lt & hi \\
\midrule
ML-Scratch N$\to$1 & 39.6 & 32.3 & 38.0 & 19.2 & 31.6 & 38.6 & 30.6 & 25.9 & 29.3 & 22.1 & 30.5 & 26.3 \\
ML-Scratch N$\to$N & 38.3 & 31.2 & 37.0 & 17.5 & 31.6 & 38.0 & 29.9 & 24.8 & 28.4 & 21.1 & 30.5 & 25.3 \\
\hline
ML-Scratch 1$\to$N & 39.1 & 23.9 & 38.5 & 20.9 & 29.3 & 28.6 & 24.6 & 31.7 & 21.2 & 17.6 & 14.5 & 19.8 \\
ML-Scratch N$\to$N & 37.2 & 23.1 & 37.8 & 20.0 & 29.1 & 27.4 & 23.1 & 30.5 & 20.3 & 16.5 & 14.6 & 19.7 \\
\bottomrule
\toprule
Lang & et & ta & ro & ps & si & ml & nl & ne & it & ar & ko & he \\
\midrule
ML-Scratch N$\to$1 & 29.1 & 20.5 & 36.3 & 16.0 & 15.4 & 19.5 & 34.5 & 17.7 & 40.1 & 51.0 & 29.2 & 39.7 \\
ML-Scratch N$\to$N & 28.3 & 19.9 & 36.6 & 15.7 & 16.2 & 19.2 & 37.6 & 20.3 & 41.9 & 44.5 & 24.1 & 40.5 \\
\hline
ML-Scratch 1$\to$N & 19.2 & 33.3 & 36.1 & 8.4 & 4.2 & 25.0 & 32.6 & 9.4 & 36.5 & 21.7 & 19.3 & 29.6 \\
ML-Scratch N$\to$N & 18.6 & 32.1 & 35.2 & 8.3 & 3.9 & 23.8 & 31.9 & 9.1 & 36.6 & 20.9 & 18.1 & 28.1 \\
\bottomrule
\toprule
Lang & tr & km & fa & vi & hr & uk & th & id & sv & pt & xh & af \\
\midrule
ML-Scratch N$\to$1 & 23.1 & 8.9 & 31.9 & 28.0 & 40.6 & 31.7 & 26.4 & 36.3 & 41.5 & 43.9 & 14.5 & 35.7 \\
ML-Scratch N$\to$N & 23.6 & 10.5 & 32.6 & 30.6 & 40.6 & 32.4 & 27.3 & 35.7 & 42.2 & 44.5 & 13.5 & 35.1 \\
\hline
ML-Scratch 1$\to$N & 22.1 & 5.0 & 18.5 & 32.5 & 32.5 & 24.4 & 36.5 & 34.7 & 38.2 & 41.9 & 4.9 & 20.3 \\
ML-Scratch N$\to$N & 21.7 & 5.0 & 18.3 & 31.9 & 31.6 & 24.5 & 36.7 & 35.4 & 38.4 & 42.0 & 8.9 & 17.6 \\
\bottomrule
\toprule
Lang & kk & ur & mk & te & sl & my & ka & gl & mr & gu & mn & az \\
\midrule
ML-Scratch N$\to$1 & 12.5 & 28.6 & 36.7 & 37.8 & 32.4 & 27.9 & 23.0 & 35.8 & 14.9 & 3.1 & 10.8 & 14.1 \\
ML-Scratch N$\to$N & 13.6 & 30.2 & 37.6 & 40.1 & 30.8 & 27.6 & 24.2 & 36.0 & 14.9 & 3.5 & 12.5 & 16.0 \\
\hline
ML-Scratch 1$\to$N & 7.9 & 24.6 & 28.3 & 41.2 & 23.4 & 35.5 & 13.5 & 28.9 & 13.9 & 3.0 & 9.2 & 8.5 \\
ML-Scratch N$\to$N & 7.9 & 24.3 & 29.5 & 41.2 & 22.6 & 36.3 & 13.2 & 28.8 & 13.8 & 3.9 & 9.1 & 7.9 \\
\bottomrule
\end{tabular}
    \caption{Multilingual Baselines over $50$ languages}
    \label{tab:ml-baselines}
\end{table*}

\begin{table*}[ht]
    \centering
    \begin{tabular}{| c|c|c|c|c|c|c|c|c|c|c|c|c |}
\toprule
Lang & de & cs & fr & ja & es & ru & pl & zh & fi & lv & lt & hi \\
\midrule
ML-FT N$\to$1 & 41.5 & 34.2 & 39.8 & 20.5 & 28.6 & 39.1 & 32.9 & 26.8 & 31.3 & 23.1 & 31.6 & 27.2 \\
ML-FT N$\to$N & 37.9 & 31.7 & 37.3 & 17.4 & 27.3 & 37.9 & 30.0 & 24.8 & 29.0 & 21.8 & 30.4 & 25.5 \\
\hline
ML-FT 1$\to$N & 38.6 & 24.5 & 38.9 & 21.8 & 29.5 & 28.7 & 24.7 & 32.4 & 21.0 & 17.9 & 14.7 & 20.0 \\
ML-FT N$\to$N & 36.8 & 23.3 & 37.4 & 20.5 & 28.6 & 27.3 & 23.1 & 31.1 & 19.7 & 16.2 & 14.4 & 18.7 \\
\bottomrule
\toprule
Lang & et & ta & ro & ps & si & ml & nl & ne & it & ar & ko & he \\
\midrule
ML-FT N$\to$1 & 30.9 & 20.9 & 38.6 & 16.2 & 17.5 & 19.9 & 38.1 & 21.1 & 43.9 & 39.1 & 21.7 & 43.5 \\
ML-FT N$\to$N & 28.4 & 19.8 & 37.0 & 15.2 & 16.1 & 18.7 & 37.7 & 19.4 & 43.3 & 41.9 & 23.3 & 42.0 \\
\hline
ML-FT 1$\to$N & 19.6 & 33.4 & 36.4 & 8.4 & 4.1 & 24.8 & 32.6 & 9.0 & 37.5 & 21.2 & 19.4 & 29.0 \\
ML-FT N$\to$N & 18.5 & 32.5 & 35.5 & 8.2 & 3.3 & 23.6 & 31.1 & 8.5 & 35.9 & 20.0 & 18.5 & 27.4 \\
\bottomrule
\toprule
Lang & tr & km & fa & vi & hr & uk & th & id & sv & pt & xh & af \\
\midrule
ML-FT N$\to$1 & 24.8 & 11.2 & 35.7 & 33.1 & 44.3 & 36.2 & 30.3 & 39.1 & 46.9 & 49.3 & 14.2 & 42.4 \\
ML-FT N$\to$N & 24.3 & 10.7 & 34.0 & 32.7 & 42.7 & 34.2 & 29.1 & 37.9 & 45.1 & 47.1 & 16.6 & 42.2 \\
\hline
ML-FT 1$\to$N & 22.1 & 6.2 & 18.3 & 32.5 & 31.9 & 24.4 & 36.0 & 34.8 & 37.8 & 41.0 & 8.9 & 20.7 \\
ML-FT N$\to$N & 21.4 & 5.7 & 18.2 & 32.0 & 30.8 & 24.1 & 35.7 & 35.1 & 38.0 & 40.8 & 11.6 & 19.6 \\
\bottomrule
\toprule
Lang & kk & ur & mk & te & sl & my & ka & gl & mr & gu & mn & az \\
\midrule
ML-FT N$\to$1 & 19.3 & 31.4 & 42.5 & 44.0 & 33.9 & 32.1 & 28.6 & 40.6 & 17.4 & 15.8 & 13.6 & 19.9 \\
ML-FT N$\to$N & 15.6 & 31.7 & 39.4 & 41.8 & 31.6 & 29.7 & 24.5 & 36.9 & 15.4 & 5.4 & 12.8 & 17.4 \\
\hline
ML-FT 1$\to$N & 6.5 & 24.6 & 27.0 & 41.0 & 22.8 & 35.4 & 12.3 & 28.0 & 13.4 & 1.9 & 8.5 & 8.1 \\
ML-FT N$\to$N & 6.9 & 22.2 & 29.0 & 39.6 & 23.1 & 36.8 & 12.3 & 28.0 & 13.1 & 1.9 & 7.7 & 8.0 \\
\bottomrule
\end{tabular}
    \caption{Multilingual Finetuning over $50$ languages}
    \label{tab:mbt50-results}
\end{table*}

\end{document}